\documentclass[sigconf]{acmart}
\AtBeginDocument{%
  \providecommand\BibTeX{{%
    \normalfont B\kern-0.5em{\scshape i\kern-0.25em b}\kern-0.8em\TeX}}}

\setcopyright{none}
\renewcommand\footnotetextcopyrightpermission[1]{} 

\usepackage{multicol}
\usepackage{epsfig}
\usepackage{amsmath}
\usepackage{multirow}
\usepackage{booktabs}
\usepackage{subfigure}
\usepackage{color}
\usepackage[switch]{lineno}
\acmConference[]{Leveraging Fine-Grained Information and Noise Decoupling for Remote Sensing Change Detection}
\begin{document}

%%
%% The "title" command has an optional parameter,
%% allowing the author to define a "short title" to be used in page headers.
\title{Leveraging Fine-Grained Information and Noise Decoupling for Remote Sensing Change Detection}

%%
%% The "author" command and its associated commands are used to define
%% the authors and their affiliations.
%% Of note is the shared affiliation of the first two authors, and the
%% "authornote" and "authornotemark" commands
%% used to denote shared contribution to the research.
% \author{Ben Trovato}
% \authornote{Both authors contributed equally to this research.}
% \email{trovato@corporation.com}
% \orcid{1234-5678-9012}
% \author{G.K.M. Tobin}
% \authornotemark[1]
% \email{webmaster@marysville-ohio.com}
% \affiliation{%
%   \institution{Institute for Clarity in Documentation}
%   \streetaddress{P.O. Box 1212}
%   \city{Dublin}
%   \state{Ohio}
%   \country{USA}
%   \postcode{43017-6221}
% }

\author{Qiangang Du}
\authornote{Equal contribution.}
\affiliation{
  \institution{Fudan University}
  \country{China}
}
\author{Jinlong Peng}
\authornotemark[1]
\affiliation{
  \institution{Tencent Youtu Lab} 
  \country{China}
}
\author{Changan Wang}
\affiliation{
  \institution{Tencent Youtu Lab} 
  \country{China}
}

\author{Xu Chen}
\affiliation{
  \institution{Tencent Youtu Lab} 
  \country{China}
}
\author{Qingdong He}
\affiliation{
  \institution{Tencent Youtu Lab} 
  \country{China}
}
\author{Wenbing Zhu}
\affiliation{
  \institution{Fudan University}
  \country{China}
}
\author{Mingmin Chi}
\affiliation{
  \institution{Fudan University}
  \country{China}
}
\author{Yabiao Wang}
\affiliation{
  \institution{Tencent Youtu Lab}
  \country{China} 
}
\author{Chengjie Wang}
\affiliation{
  \institution{Tencent Youtu Lab}
  \country{China}
}
% \affiliation{
%   \institution{Fudan University}
  
% }
% \affiliation{
%   \institution{Tencent Youtu Lab}
  
% }

% \author{Ben Trovato}
% \authornote{Both authors contributed equally to this research.}
% \email{trovato@corporation.com}
% \orcid{1234-5678-9012}
% \author{G.K.M. Tobin}
% \authornotemark[1]
% \email{webmaster@marysville-ohio.com}
% \affiliation{%
%   \institution{Institute for Clarity in Documentation}
%   \streetaddress{P.O. Box 1212}
%   \city{Dublin}
%   \state{Ohio}
%   \country{USA}
%   \postcode{43017-6221}
% }
%% You do not have to enter your paper ID

%%
%% By default, the full list of authors will be used in the page
%% headers. Often, this list is too long, and will overlap
%% other information printed in the page headers. This command allows
%% the author to define a more concise list
%% of authors' names for this purpose.
\renewcommand{\shortauthors}{Qiangang Du, et al.}

%%
%% The abstract is a short summary of the work to be presented in the
%% article.
\begin{abstract}
  Change detection aims to identify remote sense object changes by analyzing data between bitemporal image pairs. Due to the large temporal and spatial span of data collection in change detection image pairs, there are often a significant amount of task-specific and task-agnostic noise. Previous effort has focused excessively on denoising, with this goes a great deal of loss of fine-grained information. In this paper, we revisit the importance of fine-grained features in change detection and propose a series of operations for fine-grained information compensation and noise decoupling (FINO). First, the context is utilized to compensate for the fine-grained information in the feature space. Next, a shape-aware and a brightness-aware module are designed to improve the capacity for representation learning. The shape-aware module guides the backbone for more precise shape estimation, guiding the backbone network in extracting object shape features. The brightness-aware module learns a overall brightness estimation to improve the model's robustness to task-agnostic noise. Finally, a task-specific noise decoupling structure is designed as a way to improve the model's ability to separate noise interference from feature similarity. With these training schemes, our proposed method achieves new state-of-the-art (SOTA) results in multiple change detection benchmarks. The code will be made available.
\end{abstract}


%% A "teaser" image appears between the author and affiliation
%% information and the body of the document, and typically spans the
%% page.
% \begin{teaserfigure}
%   \includegraphics[width=\textwidth]{sampleteaser}
%   \caption{Seattle Mariners at Spring Training, 2010.}
%   \Description{Enjoying the baseball game from the third-base
%   seats. Ichiro Suzuki preparing to bat.}
%   \label{fig:teaser}
% \end{teaserfigure}

% \received{20 February 2007}
% \received[revised]{12 March 2009}
% \received[accepted]{5 June 2009}

%%
%% This command processes the author and affiliation and title
%% information and builds the first part of the formatted document.
\maketitle

\section{Introduction}
Change Detection (CD) is a crucial branch of Earth Remote Sensing (ERS) image analysis, which aims at identifying land surface object changes between bitemporal image pairs. It has widespread applications in detecting changes in ground buildings, land, forests and plays a key role in environmental protection, urban development. These image pairs are acquired by remote sensors in space or in the air at high altitudes over different time periods. Each pixel in the image represents a geographical area and is assigned a binary label indicating whether or not the object in that area has changed. Due to the long time span of image data acquisition, the differences between these bitemporal image pairs often contain a significant amount of noise. These noises can be categorized into task-specific and task-agnostic noise. Task-agnostic noise manifests as overall changes between image pairs, such as changes in seasonality and lighting that cause similar objects to appear differently between the bitemporal image pairs. It presents as intra-class differences. Task-specific noise, on the other hand, exhibits specific semantic changes within regions. For example, objects of different categories exhibit highly similar semantic feature descriptors, the semantic changes between different object classes are not changes of interest. These noises are often highly coupled to the relevant change in the latent space and makes it challenging to separate them. 
\begin{figure}[t]
    \centering
    \includegraphics[width=\columnwidth]{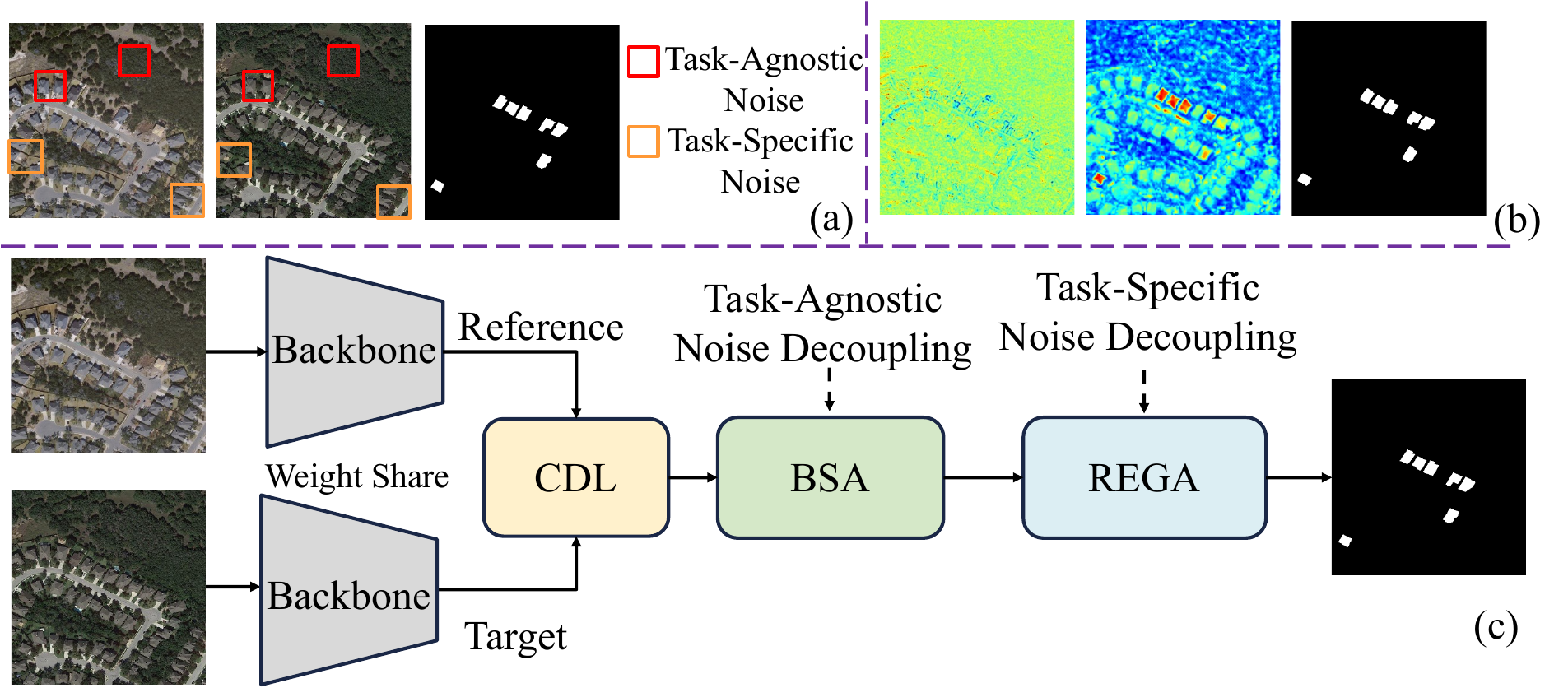}
    \caption{Conceptual illustration of proposed approach. (a) Two types of change detection pseudo-change. (b) Multi-scale fused features (far left) visualised with features learned by FINO (middle), with FINO predictions on the far right. (c) The core process of FINO. Context-dependent learning (CDL) compensates for fine-grained features, brightness-aware and shape-aware (BSA) perception to decouple task-agnostic noise, and regularization gate (REGA) decouples task-specific noise.}
    \label{fig:overiew}
\end{figure}
\begin{figure*}[htb]
\begin{center}
\includegraphics[width=0.9\linewidth]{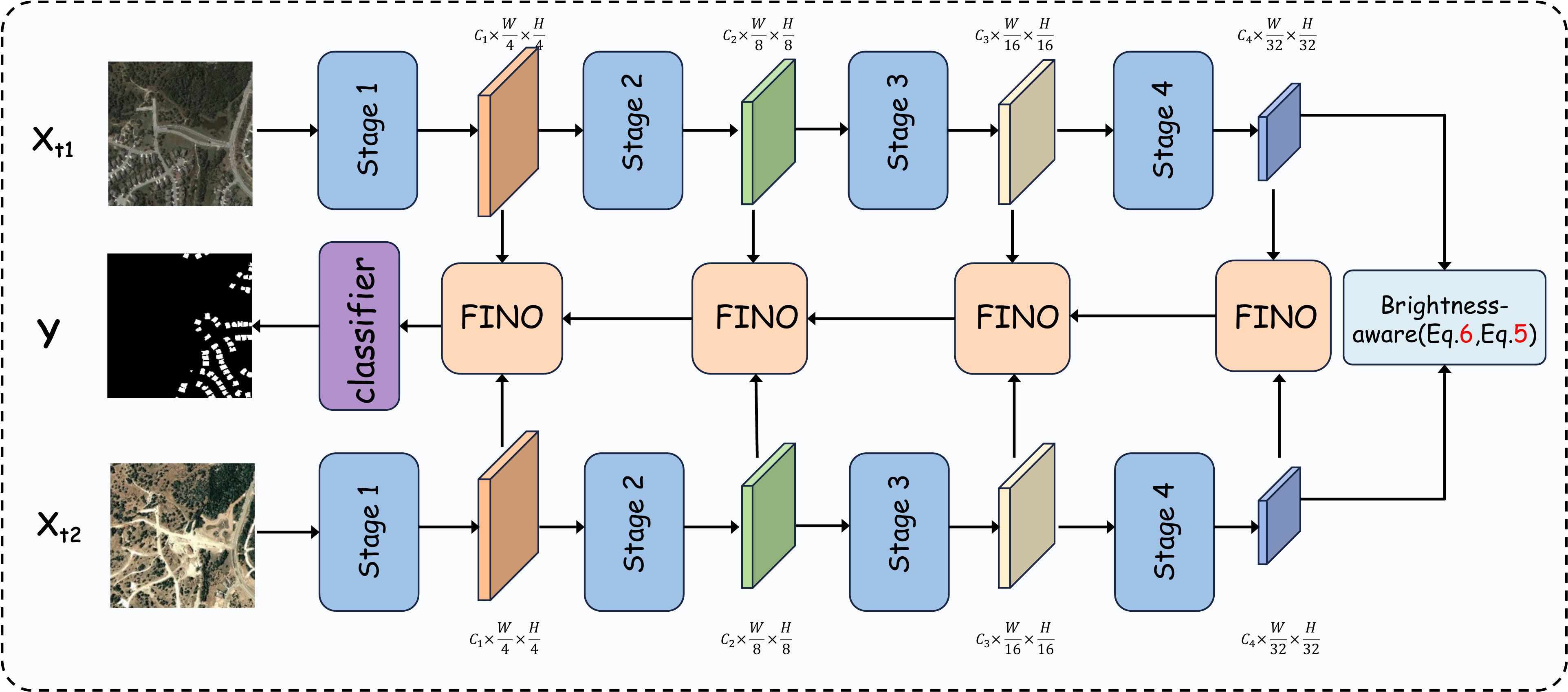}
\end{center}
    \caption{\textbf{Framework of the proposed network}. FINO consists of CDL, shape-aware module, and REGA in tandem. The attention-based CDL adequately compensates for fine-grained information. The shape-aware module decouples the task-agnostic noise and guides the model to learn the object shape representation. REGA decouples the task-specific noise. }
    \label{fig:framwork}
\end{figure*}

Daudt et al.~\cite{ieee:Sia-CD} first proposed a Siamese networks based Deep Learning (DL) for change detection, then DL has achieved excellent results in change detection. The current mainstream approach is based on siamese backbones to extract features separately and detect changed objects by
the difference information. Recent efforts have focused on designing specialized networks to filter out the mentioned noise. However, by overly focusing on denoising, these methods inevitably suffer from the loss of some crucial detail information, which is essential for change detection, such as location, scale and more. These approaches address this issue by multi-scale features fusion~\cite{ussfc, DSIFN-CD, chen2023sarasnet, levir-cd, BIT, ChangeFomer}, feature interaction~\cite{chen2023sarasnet,arxiv:changer, 22tgars:RSCD, wang2023apd} or skip connections~\cite{2018:ECCV-changenet, tgars:dessn} to compensate for fine-grained information. Nevertheless, these methods inevitably lead to the problems of information redundancy and noise regeneration, as shown in Fig.~\ref{fig:overiew}(b). Furthermore, as the complexity and depth of the network model increasing, it will lead to overfitting.

In summary of the above review, existing methods overly prioritize denoising and ignore some key issues, as shown in Fig.~\ref{fig:overiew}(a). \textbf{Firstly}, visual understanding belongs to fine-grained task. Fine-grained details are crucial for identifying the aforementioned pseudo-change noise. However, previous works have largely neglected this aspect and instead solely focused on noise elimination. \textbf{Secondly}, global features in change detection primarily manifest in overall descriptions like illumination and seasonality. Simply pursuing long distance feature learning does not significantly enhance detection performance~\cite{BIT, ChangeFomer}. Change detection exhibits a high degree of local contextual similarity. There is a high probability that an object is surrounded by objects of its own kind. \textbf{Finally}, the noise in change detection consists partly of task-specific and task-agnostic components. Treating them as a single entity for denoising while learning features can be disastrous. Extreme data imbalance severely hampers noise suppression, and few works have addressed this issue. 

In this paper, we claim that current research in change detection lacks sufficient utilization of fine-grained information and the ability to differentiate noise. We propose a series of operations for \textbf{F}ine-grained \textbf{I}nformation compensation and \textbf{N}oise dec\textbf{O}upling (FINO) to improve change detection representation learning, as shown in Fig.~\ref{fig:overiew}(c). First, to acquire local features and compensation for fine-grained information, we first employ a context-dependent learning (CDL) module based region attention mechanism to learn contextual semantic features. This approach allows us to gradually reconstruct fine-grained features by using high-level semantic features to guide the reconstruction of lower-level features. Following it, based on alignment operations, we propose a brightness-aware and shape-aware (BSA) module to guide the model in learning object shape descriptions and enhance the model robustness to task-agnostic noise. Finally, we introduce a regularization gate (REGA) structure to decouple task-specific noise. On one hand, the gate structure prevents the noise reintroducted during denoising and improves the confidence of interest object changes. On the other hand, through regularization gating, we deactivate a portion of neurons during gradient computation to enhance the model robustness to task-specific noise.

In summary, the main contributions of this paper are as follows:
\begin{itemize}
    \item 1. We rethink the importance of fine-grained information in change detection and elucidate the nature of pseudo-change noise. A context-dependent learning structure based regional attention is proposed to compensate for fine-grained information.
    \item 2. We propose brightness-aware and shape-aware modules to enhance the representation learning of shapes and improve model robustness against task-agnostic noise.
    \item 3. We introduce a regularization gate structure to decouple task-specific noise which helps mitigate the impact of extreme data imbalance on the model learning in change detection.
    \item 4. We achieve SOTA performance on multiple change detection benchmarks. Compared to the SOTA methods, our method improves F1 by 0.60\% and IoU by 1.03\% on the LEVIR-CD~\cite{levir-cd} dataset. On the WHU~\cite{WHU} dataset, the F1 is improved by 0.86\% and IoU by 4.97\%. On the DSIFN-CD~\cite{DSIFN-CD} dataset, our approach leads to a 1.48\% improvement in F1 and a 2.39\% increase in IoU. Finally, on the CDD~\cite{CDD2018Lebe} dataset, we saw an increase of 1.17\% in F1 and 2.25\% in IoU. 
\end{itemize}
\section{Related Work}
\subsection{Siamese Network for CD}
The task of change detection presents unique challenges compared to image classification or segmentation tasks due to its sensitivity to both the bitemporal image features themselves and the difference features that represent whether there are changes. The input to change detection is an image pair consisting of the current temporal image and the past temporal image, making Siamese networks a preferred choice due to their ability to map the respective images into a new space with fewer parameters but the same structure. Most of the existing works~\cite{FCN,levir-cd, 22tgars:RSCD, arxiv:changer, DSIFN-CD, peng2021siamrcr, tgars:dessn, wang2023apd} use Siamese networks as a backbone. Daudt et al.~\cite{ieee:Sia-CD} first proposed Siamese architectures for CD and compared the Early Fusion (EF) and Siamese, proving that last fusion (LF) such as Siamese is superior to EF. However, the inevitable drawback of these methods is that the downsampling causes the loss of detailed information about the data as the depth of the neural network increases. Ziming Li et al.~\cite{tgars:darn, tgars:dessn} bridge the high-level semantic information and the low-level detailed information through a skip connection structure, thus ensuring that the extracted changed features contain both high-level features and local detailed information. Zhenglai Li et al.~\cite{22tgars:RSCD} use several repetitive multi-scale change information interaction structures to extract change features. Although the above methods enhance the interaction and fusion of information between different branches of Siamese level, they result in information redundancy and regenerate a large amount of noise. It's because that the irrelevant change noises are highly adherent to the changed features on the feature space, these methods that directly adopt raw space features for fusion or interaction would regenerate noise to downstream change detection.

In this paper we refine the change detection noise by decoupling task-specific and task-agnostic noise in REGA and BSA to learn change features more accurately.
\subsection{Information Aggregation For CD}
Change detection in images often relies on the detailed features of the original image to provide valuable information about the changes. However, existing methods, such as feature difference-based change detection, often fail to capture critical information. To address these limitations, information fusion methods have been proposed, broadly including multi-scale information aggregation~\cite{tgars:dessn, 22tgars:RSCD}, difference feature and image feature information aggregation~\cite{22tgars:ssan}, and a combination of both~\cite{tgars:darn}. Unfortunately, these methods tend to introduce noise into the change information. To mitigate this problem, researchers have used methods such as LSTM and RNN to eliminate noise and improve the extraction of global semantic information~\cite{2016IGARSS:RNN,2017CVPR:LSTM,2019TGARS:LSSTF}. Bai~\cite{22TGARS:ERCN} and Yang~\cite{22TGARS:IRCNN} have proposed a change detection method using irregular RNN twin networks that combine Siamese network and RNN. However, these methods suffer from complex structures, high training difficulty, and a high risk of overfitting due to the long-term dependencies of RNN and the large number of LSTM parameters. 

In this paper we mitigate the above problem by compensating fine-grained features through CDL.
\section{Approach}
\subsection{Overall Framework}
In this section we explain our proposed method which addresses the problem of loss of fine-grained information during network downsampling and the difficulty of removing irrelevant change noise in the change detection problem. The framework of our methods is shown in Fig. ~\ref{fig:framwork}.

Denote the pair of bitemproal images as $\{X^{t1}, X^{t2}\}\in \mathbb{R}^{C\times H\times W}$, which indicates the past-image and the post-image respectively. And the labelled data is denoted by $Y\in \{0,1\}$.  We define change detection as the process of determining the distance between two distributions, which is considered to have changed when the distance exceeds a certain threshold, i.e. the information represented by the feature has changed. The process can be defined as follows: 
\begin{align}
    p &= P(\mathcal D(\mathcal F(X^{t1}),\mathcal F(X^{t2}))|X^{t1}, X^{t2}) \\
    y &=
    \begin{cases}
        0,  \quad p\leq T \\
        1,  \quad else \\
    \end{cases}
\end{align}
where $P$ denotes the probabilistic classifier, $p$ denotes the derived probability, $\mathcal D$ denotes the distance of the distribution, $\mathcal F$ denotes the feature mapping function and $T$ denotes the threshold. Our approach focuses on the noise decoupling of $\mathcal D$ process and the features adequate learning among $\mathcal F$ process.

Our proposed architecture consists of one backbone for feature extraction, three main components for fine-grained information compensation and noise decoupling, one segmentation head as classifier. Firstly the feature extraction backbone is based on weight-shared Siamese ResNet~\cite{he2015resnet}. Secondly our core method of Fine-grained Information compensation and Noise decOupling (FINO) consists of context-dependent learning (CDL) module, brightness-aware and shape-aware learning (BSA) and regularization gate (REGA) structure. The segmentation head consists of a 1 $\times$ 1 convolution and three 3 $\times$ 3 convolutions in series. 

At the beginning, a Siamese network is adopted to map the input images $X^{t1}, X^{t2}$ to low resolution feature tensor. We use features from four stages within the backbone network as representations for downstream change detection. Specifically, we utilize features at resolutions of 1/4, 1/8, 1/16, and 1/32 of the original image resolution. Let $\{X_i^{t1},X_i^{t2}\}$ denotes the feature maps extracted by the $i$th stage of the ResNet~\cite{he2015resnet} backbone and $C_{i}$ is the feature difference of $X_i^{t1}$ and $X_i^{t2}$.
\begin{figure}[htb]
    \centering
    \includegraphics[width=0.9\columnwidth]{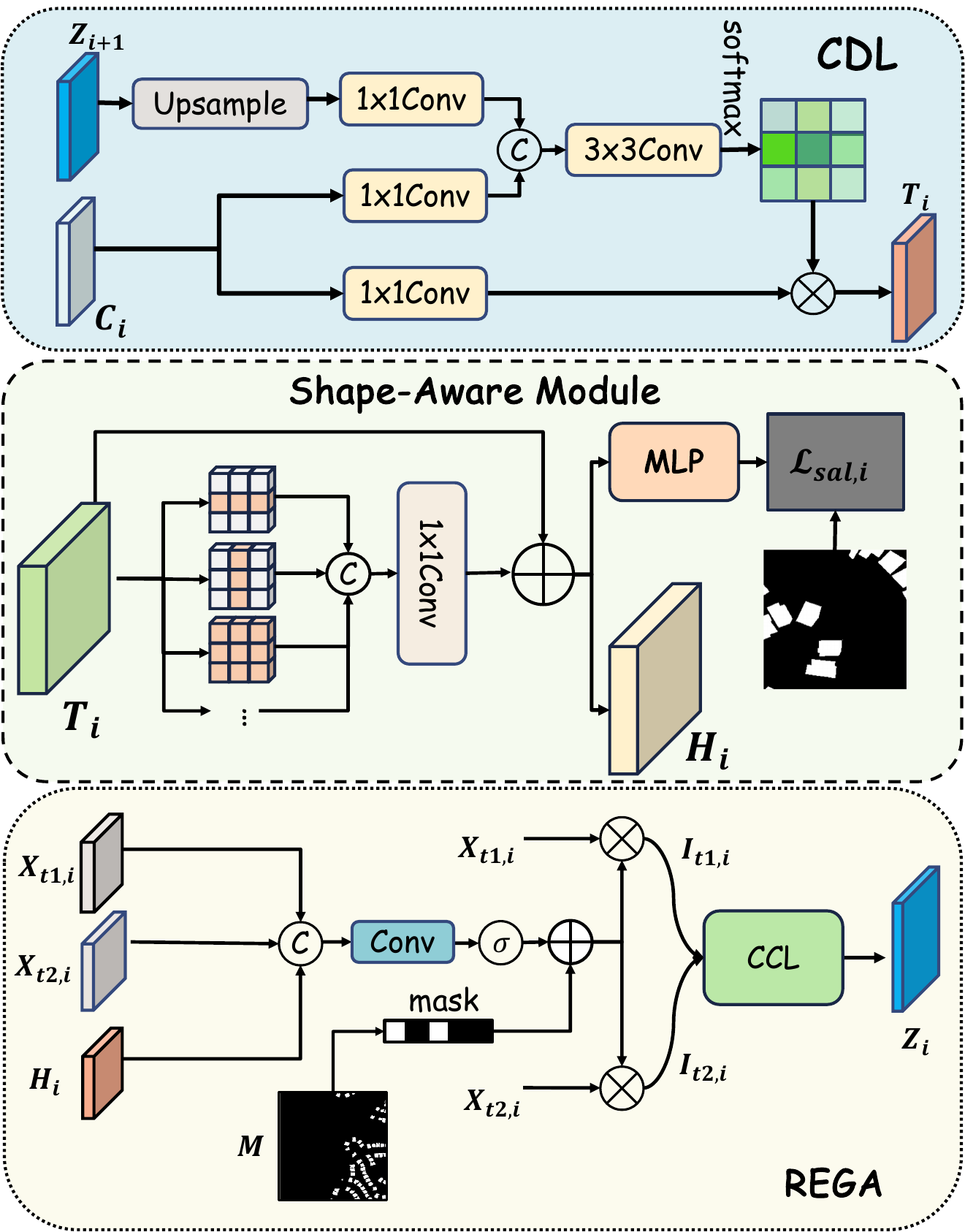}
    \caption{FINO Detail Structure. FINO consists of CDL, shape-aware module and REGA. First, CDL compensates the fine-grained information in $C_i$ through $Z_{i+1}$ to obtain the compensated feature $T_i$. Then, the shape-aware module learns the shape to obtain $M_i$ and enriched feature $H_i$. In REGA, $H_i$ and the bitemporal features $X_{t_1,i} X_{t_2,i}$ pass through a gated structure to obtain the change feature $Z_i$ of the current layer, where $M_i$ is used as a regularization term for the gated structure to improve the robustness to pseudo changes.}
    \label{fig:fino}
    \vspace{-6mm}
\end{figure}
\subsection{Context-Dependent Learning}\label{CIFR}
We claim that the importance of local descriptors within regions far outweighs global descriptors in change detection. Global descriptors mainly manifest as color and brightness changes caused by factors like seasonality and lighting, which can be learned using lightweight global networks. On the other hand, local region features are crucial for decoupling noise between paired images. Based on this understanding, we propose a region-wise attention mechanism that leverages rich semantic features from higher-level network layers to guide lower-level networks in compensating for fine-grained features. In order to obtain the dependencies between image regions, we propose CDL to compensate the fine-grained information which lost during the extraction of high-level semantic features. The CDL consists of multiple region attention modules that are connected in a cascading manner. The architecture is shown in the Fig.~\ref{fig:fino}. To guide the low-level features and fuse the high-level low-noise with the low-level high-noise features, $C_i$ and $Z_{i+1}$ are input to CDL and output $M_i$, where $M_i$ denotes the changed feature of $i$th layer and contains both semantic features and fine-grained features. 

As the higher level features have stronger semantic information with lower resolution, whereas lower-level network features possess detailed information but at a higher resolution. To apply attention mechanisms efficiently on high-resolution features at minimal cost, we use a localized region attention mechanism instead of the traditional matrix multiplication-based attention mechanism works~\cite{tgars:dessn, BIT, ChangeFomer, noman2023scratchformer, peng2020chained}. This localized attention mechanism focuses on smaller regions within the feature maps, allowing for effective utilization of detailed information without excessive computational overhead. We correct the change features of the current layer by an attention of $A_i$ as follows:
\begin{align}
    A_i(C_i, Z_{i+1}) &= Softmax(\frac{flatten(\varphi_1(C_i,Z_{i+1}))}{\sqrt{d_k}}) \\
    T_i &= A_i \varphi_2(C_i)
\end{align}
where $d_k$ represents the feature dimension, $C_i$ denotes the difference feature of the Stage-$i-th$ layer in Fig.~\ref{fig:fino}, $T_i$ denotes the region-aware contextual features learned by CDL and $\varphi_1, \varphi_2 $ denotes the convolution function. 

Since change is often of a localised nature, excessive focus on global attention is not effective. CDL is a local attention that differs from the normal attention. Self-attention uses a attention vector multiplication, while we use element multiplication to compute the attention map. 
\subsection{Brightness-Aware and Shape-Aware Learning}\label{GDFA}
Typically, due to the large time span of data acquisition for change detection image pairs, objects between image pairs tend to have task agnostic noise. These differences are unpredictable for the change detection task itself and represent global image characteristics. We aim to decouple these noises using brightness-aware and shape-aware modules. And brightness-aware module is only applied at the last layer of the backbone.

\textbf{Brightness-aware learning}. Firstly, the brightness-aware module consists of two contrastive learning components: one for global brightness feature contrastive learning and another for region-level semantic feature alignment. We aim for the backbone network's features to possess robustness against brightness interference in their global features. Therefore, we use learnable global pooling to obtain global features for the paired images and employ contrastive learning to guide the backbone network's learning. The global contrast learning loss is calculated as follows:
\begin{equation}
    \mathcal L_{gcl} = 1 - \left \langle \boldsymbol{GAP}(X_{t1}), \boldsymbol{GAP}(X_{t2})\right \rangle
\end{equation}
where $\left \langle \_, \_\right \rangle$ denotes the cosine similarity, $\boldsymbol{GAP}$ indicates the AdaptiveAvgPooling.

Additionally, we aim to minimize the differences between objects in image pairs caused by factors like lighting through feature extraction by the backbone network. Therefore, the loss computation for region-level semantic feature alignment is defined as follows:
\begin{align}
    \mathcal{L}_{rcl} &=  -\frac{1}{N}\sum_k^N y_k\log s_k + (1-y_k)\log(1-s_k) \\
    s_k &= \left \langle X_{t1,4,k}, X_{t2,4,k}\right \rangle
\end{align}
where $X_{t1,4,k}, X_{t2,4,k}$ indicates the region feature of forth stage, $k$ is the pixel index in space $N=H\times W$, $\left \langle \_, \_\right \rangle$ denotes the cosine similarity, $y_k$ is the label.

\textbf{Shape-aware learning}. To refine boundary and expand the receptive field, we use a shape-aware module that consists of multiple asymmetric convolution for anti-aliasing of the edge, shown as in Fig.~\ref{fig:fino}. In our approach, we use seven convolution kernels, including $1\times1, 1\times3, 3\times1, 3\times3, 1\times5,5\times1, 5\times5$. We use Eq.~\ref{eq:shape-aware} to represent the learning of the shape-aware module:
\begin{align}\label{eq:shape-aware}
    H_i &= T_i + W_0 Asym(T_i) \\
    M_i &= MLP(H_i)
\end{align}
where $Asym$ indicates the asymmetric convolutions, $W_0$ denotes the weight of the $1\times1$ convolution and $MLP$ represents a multilayer perceptron mapping. 

Besides, we inject additional supervised objectives in the shape-aware module to ensure the accuracy of estimating shapes. We apply the binary cross-entropy as loss:
\begin{equation}
    \mathcal L_{sal} = -\sum_{i=1}^{4}\frac{1}{N}\sum_k^N \bar{y_k}\log M_{i,k} + (1-y_k)\log(1-M_{i,k})
\end{equation}
where $k$ is the pixel index, $M_{i,k}$ is the $i$-th stage layer shape mask, $y_k$ is the label.

Additionally, $M_{i}$ is fed into the regularization gate structure to guide decouple task-specific noise.
% And we regularise the gating weights by a mask discretising the output of BEM. After that, we get the real changed representations that control the flow of data to the downstream network structures. The gate structure receives the inputs from each stage of the ResNet and the outputs from the SAM, then generates a spatial data switch that based on the learnable weights. It adaptively determines whether the current feature is a relevant change and control the data flows to the downstream CD network structure.
\begin{table*}[h]
\scriptsize
    \centering
    \begin{tabular}[]{c|c|cccc|cccc|cccc|cccc}
        \toprule[2pt]
        \multirow{2}*{Methods} & \multirow{2}*{Backbone} & \multicolumn{4}{|c|}{LEVIR-CD~\cite{levir-cd}} & \multicolumn{4}{|c|}{WHU~\cite{WHU}} & \multicolumn{4}{|c|}{DSIFN-CD~\cite{DSIFN-CD}} & \multicolumn{4}{|c}{CDD~\cite{CDD2018Lebe}} \\
        % \cline{3-18}
        ~ & ~ & Pre.(\%) & Rec(\%) & F1(\%) & IoU(\%) & Pre(\%) & Rec(\%) & F1(\%) & IoU(\%) & Pre(\%) & Rec.(\%) &  F1(\%) & IoU(\%) & Pre(\%) & Rec(\%) &  F1(\%) & IoU(\%)\\
        \midrule
        FC-Siam-diff~\cite{FCN} & U-Net & 89.53 & 83.31 & 86.31 & 75.92 & 47.33 & 77.66 & 58.81 & 41.66 & 59.67 & 65.71 & 62.54 & 45.50 & 93.65 & 54.32 & 68.76 & 47.74  \\
        SNUNet~\cite{SNUNet} & U-Net++ & 89.18 & 87.17 &  88.16 & 78.83 & 85.60 & 81.49 & 83.50 & 71.67 & 60.60 & 72.89 & 66.18 & 49.45 & 95.19 & 92.51 & 93.83 & 88.38 \\
        USSFC-Net~\cite{ussfc} & U-Net & 89.70 & \textcolor{red}{92.42} & 91.04 & 83.55 & 89.96 & \textcolor{blue}{94.56} & 92.20 & 85.54 & 63.73 & 76.32 & 69.47 & 53.21 & 93.35 & 96.08 & 94.74 & 90.02 \\
        IFNet~~\cite{zhang2020deeply} & VGG16 & $ \textcolor{red}{94.02}$ & 82.93 &  88.13 & 78.77 & 96.91 & 73.19 & 83.40 & 71.52 &  67.86 & 53.94 & 60.10 & 42.96 & 97.64 & 96.35 & 96.99 & 88.94 \\ 
        DASNet~\cite{9259045Dasnet} & VGG16 & 90.60 & 91.38 & 90.99 & 83.47 & 88.23 & 84.62 & 86.39 & 76.04 &  60.10 & 56.53 & 58.26 & 41.10 & 92.50 & 91.40 & 91.90 & - \\
        DTCDSCN~\cite{DTCDSCN} & ResNet34 & 88.53 & 86.83 &  87.67 & 78.05 & 63.92 & 82.30 & 71.95 & 56.19 & 53.87 & 77.99 & 63.72 & 46.76  & - & - & - & - \\
        TinyCD~\cite{codegoni2022tinycd} & EfficientNet & 82.68 & 89.47 &91.05 &83.57 & 91.72 & 91.76 & 91.74 & 84.74 & 51.48 & 77.66 & 61.92 & 44.84 &  94.78 & 93.24 & 94.00 & 88.68 \\
        ScratchFormer~\cite{noman2023scratchformer} & ResNet50 & 92.70 & 89.06 & 90.84 & 83.22 & 88.76 & 84.13 & 86.39 & 76.03 & 88.76 & 84.14 & 84.65 & 73.40 & 96.29 & 96.00 & 96.14 & 92.57 \\
        SARAS-Net~\cite{chen2023sarasnet} & ResNet50 & 91.97 & \textcolor{blue}{91.85} & 91.91 & 84.95 & 88.41 & 85.81 & 87.09 & 77.14 & 67.65 & 67.51 & 67.58 & 51.04 & \textcolor{blue}{97.76} & 97.23 & \textcolor{blue}{97.49} & \textcolor{blue}{95.11} \\ 
        STANet~\cite{ma2023stnet} & ResNet18 & 83.81 & 91.00 &  87.26 & 70.40 & 79.37 & 85.50 & 82.32 & 69.95 & 67.71 & 61.68 & 64.56 & 47.66 & 95.17 &  92.88 & 94.01 & 87.98 \\
        BiT~\cite{BIT} & ResNet18 & 89.24 & 89.37 &  89.31 & 80.68 & 86.64 & 81.48 & 83.98 & 72.39 & 68.36 & 70.18 & 69.26 & 52.97 & 94.57 & 88.18 & 91.26 & 83.93 \\
        Changer~\cite{arxiv:changer} & ResNet18 & 92.86 & 90.78 & \textcolor{blue}{91.81} &\textcolor{blue}{84.86} & \textcolor{red}{95.29} & 89.90 & \textcolor{blue}{92.49} & \textcolor{blue}{85.99} & 88.67 & 82.20 & 85.31 & 74.39 & 88.67 & 82.2 & 85.31 & 74.39 \\ 
        STNet~\cite{ma2023stnet} & ResNet18 & 92.06 &89.03 &90.52 &82.09 & 87.84& 87.08& 87.46&77.72 & - & - & - & -  & - & - & - & - \\
        APD~\cite{wang2023apd} & ResNet18 & 92.81 & 90.64 & 91.71 & 84.69 & 95.74 & 86.94 & 91.13 & 83.70 & \textcolor{blue}{89.39} & \textcolor{blue}{86.40} & \textcolor{blue}{87.87} & \textcolor{blue}{78.36} & 96.36 & 96.01 & 96.20 & 92.67\\
        CANet~\cite{CANet_zhou} & ResNet18 & 92.52& 90.06& 91.27& 83.95& - & - & - & - & - & - & - & - & 96.82 & \textcolor{blue}{97.29} & 97.05 & 94.27 \\ %\hline
        \midrule[0.2pt]
        % Siam-ResNet & {\large 87.26 & 86.70&  87.02 & 77.03 & - & - & - & - & 52.36 & 55.13& 53.71 & 36.71 &  &  &  &  \\
        FINO(Ours)& ResNet18 &  \textcolor{blue}{93.70} & 91.15 & $\textcolor{red}{92.41}$ & $\textcolor{red}{85.89}$ & $\textcolor{blue}{95.27}$ & $\textcolor{red}{97.26}$ & $\textcolor{red}{93.35}$ & $\textcolor{red}{90.96}$ & $\textcolor{red}{91.34}$ & $\textcolor{red}{96.01}$ & $\textcolor{red}{89.35}$ & $\textcolor{red}{80.75}$ & \textcolor{red}{98.76} & \textcolor{red}{98.57} & \textcolor{red}{98.66} & \textcolor{red}{97.36} \\
        \bottomrule[2pt]
    \end{tabular}
    \caption{Quantitative results on LEVIR-CD dataset, WHU-CD dataset DSIFN-CD dataset and CDD dataset. The best values are in red and the second are in blue. We add references of these models in the supplementary material.}
    \label{tab:compasion}
\end{table*}
\subsection{Regularization Gate Structure}
We assume that the shape-aware module has obtained highly confident shape features. We use a gating mechanism to selectively enhance foreground features based on this. The specific structure is illustrated in Fig.~\ref{fig:fino}. 

Firstly, REGA uses a $1\times 1$ convolution to aggregate features $\{X_{t1,i}, X_{t2,i}\}$ from paired images and shape features $H_i$ learned by asymmetric convolution. Then it passes through the gating mechanism to obtain gating weights. We use sigmoid as the gate. Furthermore, we use the shape mask $M_i$ predicted by the shape-aware module to regularize the gating weights, deactivating neurons in regions affected by pseudo-change noise. It enhances the model robustness to task-specific noise. Finally, a channel attention structure is used for change feature prediction.

The gate function can be defined as Equation~\ref{gate function}.
\begin{align}\label{gate function}
    G_i = \sigma(\varphi(CAT(X_{t1,i}, X_{t2,i}, H_i))) + M_i
\end{align}%
\begin{equation}
    I_{t,i} = X_{t,i} \otimes G_i
\end{equation}
Where $X_{t,i}\in\{{X_{t1,i}, X_{t2,i}}\}$, $I_{t,i}\in\{I_{t1,i}, I_{t2,i}\}$, $i$ denotes the layer index and $i\in\{1,2,3,4\}$, $\sigma$ represents the activation function sigmoid, $CAT$ indicates concatenation among features along channel dimension, $\otimes$ indicates spatial-wise multiplication.

\textbf{Change Characteristics Learning.}\label{DiConv}
Existing methods use the distance between two features as the basis for change detection, which lacks interpretability and may miss certain factual change information. As described earlier, we define the procedure of change detection as learning the distance between two distributions. Therefore, we adopt a change characteristics learning module (CCL) based on channel attention to obtain the final change features. 
% \begin{figure}[htb]
%     \centering
%     \includegraphics[width=0.8\columnwidth]{figures/ddm.pdf}
%     \caption{Change characteristics learning module (CCL).}
%     \label{fig:ddm}
% \end{figure}

We use the equation~\ref{diconv-eq1} and~\ref{diconv-equation} to refer the above operations.
\begin{equation}\label{diconv-eq1}
    d_i = |w\cdot I_i^{t1} - w \cdot I_i^{t2}|
\end{equation}
\begin{equation}\label{diconv-equation}
    Z_i = d_i\cdot \mathit MLP(MaxPool(d_i)+AvgPool(d_i))
\end{equation}
where $w$ denotes the weight of a $1\times1$ convolution that maps the features of bitemproal images to the same distribution space. We then obtain a pixel-by-pixel distribution of distances and input them to the segmentation head to get the prediction.

\subsection{Loss}
The overall loss function $\mathcal{L}$ consists of change detection loss and supervisory loss:
\begin{align}
    \mathcal L = \mathcal{L}_{cd}+\mathcal L_{sal} + \lambda\mathcal L_{gcl} + \lambda\mathcal L_{rcl}
\end{align}
where $\lambda$ indicates a loss weight hyperparameter. We adopt binary cross-entropy for $\mathcal{L}_{cd}$ and $\mathcal{L}_{sal}$. Based on task relevance, the change detection loss $\mathcal{L}_{cd}$ and shape-aware learning loss $\mathcal L_{sal}$ are the primary losses. The global contrast learning loss $\mathcal L_{gcl}$ and region-level alignment loss $\mathcal L_{rcl}$ are auxiliary losses, with the weight $\lambda$ set to 0.1. 
\section{Experiment}
\subsection{Settings}
\textbf{Datasets:} In this paper we use four CD datasets, named LEVIR-CD~\cite{levir-cd}, WHU~\cite{WHU}, DSIFN-CD~\cite{DSIFN-CD}, CDD~\cite{CDD2018Lebe}. 
We obtain 445/64/128 pairs of patches with size $1024^2$ for training/validation/test in LEVIR-CD datasets. We crop the one pair of WHU-CD images to non-overlaped patches with size of $512^2$. We obtain 1344/192/384 pairs of patches for training/validation/test.  
DSIFN-CD is cropped to $256^2$ and there are 14400/1360/192 pairs in training/validation/test datasets. The CDD dataset obtain 10000/3000/3000 pairs with size of $256^2$ for training/validation/test datasets.

\textbf{Implementation Details:}
Our approach is implemented by Pytorch and trained on two GeForce RTX 1080Ti GPUs with 11GB VRAM. The model is trained for 500 epoches with two GPUs, where there are two pairs of images on each GPU. We use ResNet18 as the backbone network in the following experiments. We employ data augmentation techniques, including random rotation, random cropping, brightness variation, and random flipping. The AdamW is used as the optimizer with a $\it{poly}$ learning rate policy, where the learning rate is set to 0.001 with $\gamma=0.9$, $weight\_decay=0.0001$ and policy was step.

\textbf{Metrics:}
In change detection, precision reflects resistance to pseudo-change (e.g. color changes of buildings) noise, recall measures sensitivity to actual changes, and F1-score offers a comprehensive evaluation. IoU assesses shape analysis and change object localization. In this context, the primary focus centers on comparing F1-score and IoU. 

\begin{figure*}[ht]
\begin{center}
\includegraphics[width=\linewidth]{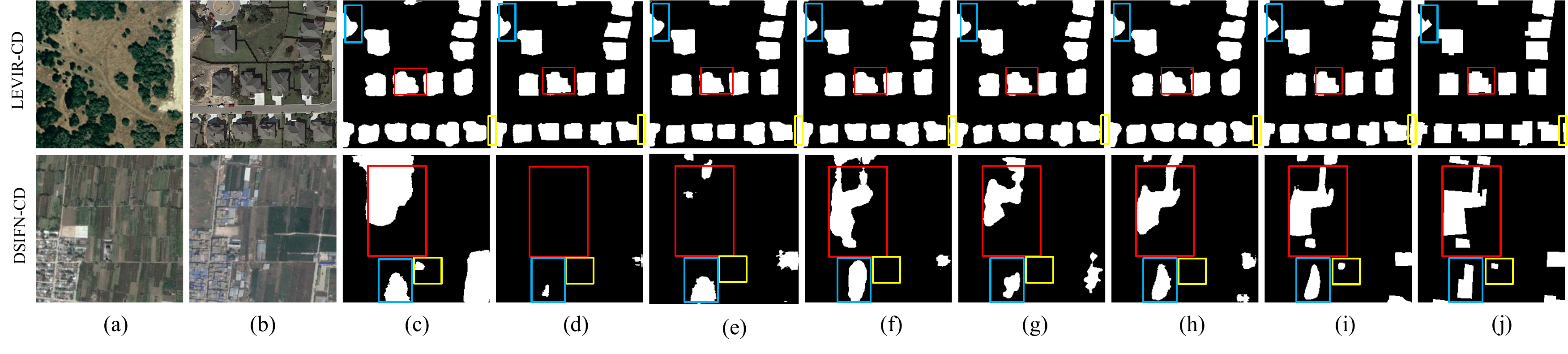}
\end{center}
    \caption{ Qualitative results of different CD methods on LEVIR-CD. (a) $T_1$ image. (b) $T_2$ image. (c) FC-EF. (d) FC-Siam-Diff. (e) FC-Siam-Conc. (f) DTCDSCN. (g) BIT. (h) ChangeFormer. (i) FINO(ours). (j) Ground truth. where the first row shows the comparison of detection results for dense objects and the second row shows the comparison of detection results for multi-scale objects. The yellow box is the detection effect of small objects, the red box is the detection effect of large objects, and the blue box is the detection effect of object edges. }
    \label{fig:vis_quality}
\end{figure*}
\subsection{Comparison With State-of-the-Art Methods} \label{sota}
\textbf{Comparison of Metrics:} In order to demonstrate the effectiveness of our method and compare it with existing excellent CD methods, we conducted comparative experiments with several methods on different datasets. The specific results are shown in Table~\ref{tab:compasion}, it can be seen that FINO outperforms other methods and has a significant margin on those datasets. Our method far outperforms the other methods of F1 and IoU on all the four datasets.

% The specific results are shown in the Table~\ref{tab:enhanceEX}. The Table~\ref{tab:enhanceEX} visualizes the considerable improvement of our method over methods such as feature fusion. On both F1 and IoU performing better than DARNet~\cite{tgars:darn} which is the the SOTA, where F1 is 0.43 higher than DARNet and IoU is 0.83 higher than DARNet. 
\begin{table}[t]
\footnotesize
    \centering
    % \resizebox{0.9\linewidth}{!}{
        \begin{tabular}[]{c|cccc}
            \toprule[1.5pt]
            Methods & Pre.(\%) & Rec.(\%) & F1(\%) & IoU(\%) \\
            \midrule
            RSCD~\cite{22tgars:RSCD}  & 92.15 & 89.73 & 90.93 & 83.36 \\
            D-TNet18~\cite{dtnet} & 89.13 & 88.58 & 88.85 & 79.94 \\
            DARNet~\cite{tgars:darn} & 92.67 & \textcolor{red}{91.31} & 91.98 & 85.16 \\
            ChangeFormer~\cite{ChangeFomer} & 92.05 & 88.80 & 90.40 &82.48 \\
            DESSN~\cite{tgars:dessn}  & 90.99 & 91.73 & 91.36 & - \\
            FINO(Ours) & \textcolor{red}{93.70} & 91.15 & \textcolor{red}{92.41} & \textcolor{red}{85.89} \\
            \bottomrule[1.5pt]
        \end{tabular}
    % }
    \caption{Comparison experiments of our method with other direct feature enhancement or feature fusion methods on LEVIR-CD dataset. }
    \vspace{-5mm}
    \label{tab:enhanceEX}
\end{table}

\textbf{Comparison of Detection Results:}
In order to demonstrate the superiority of the proposed method in detection results, we conducted a visual analysis on LEVIR-CD and DSIFN-CD. The visual analysis of experimental results are shown in Fig.~\ref{fig:vis_quality}, where the detection results on LEVIR-CD are exhibited in the first row of Fig.~\ref{fig:vis_quality} (c)-(j) and the detection results on DSIFN-CD are exhibited in the second row of Fig.~\ref{fig:vis_quality} (c)-(j). In general, FINO performs better than other methods, with smoother and more regular detected edges. For example, as shown in the blue box in Fig.~\ref{fig:vis_quality}(c)-(j), the predictions of FINO are closer to the GT and has smoother and more regular edges, which is more in line with the shape of the building. And FINO has the same sensitivity to large target changes as to small target changes, shown in the red and yellow boxes in Fig.~\ref{fig:vis_quality}(c)-(j). For example, as can be seen from the yellow box, other methods cannot detect this part of the changed object, while FINO is able to not only detect this part of the change, but its predicted label is very close to the GT and its prediction results are more complete and accurate. On the other hand, our method FINO has few errors, as can be observed in the red box.
\begin{figure*}[htb]
\begin{center}
\includegraphics[width=\linewidth]{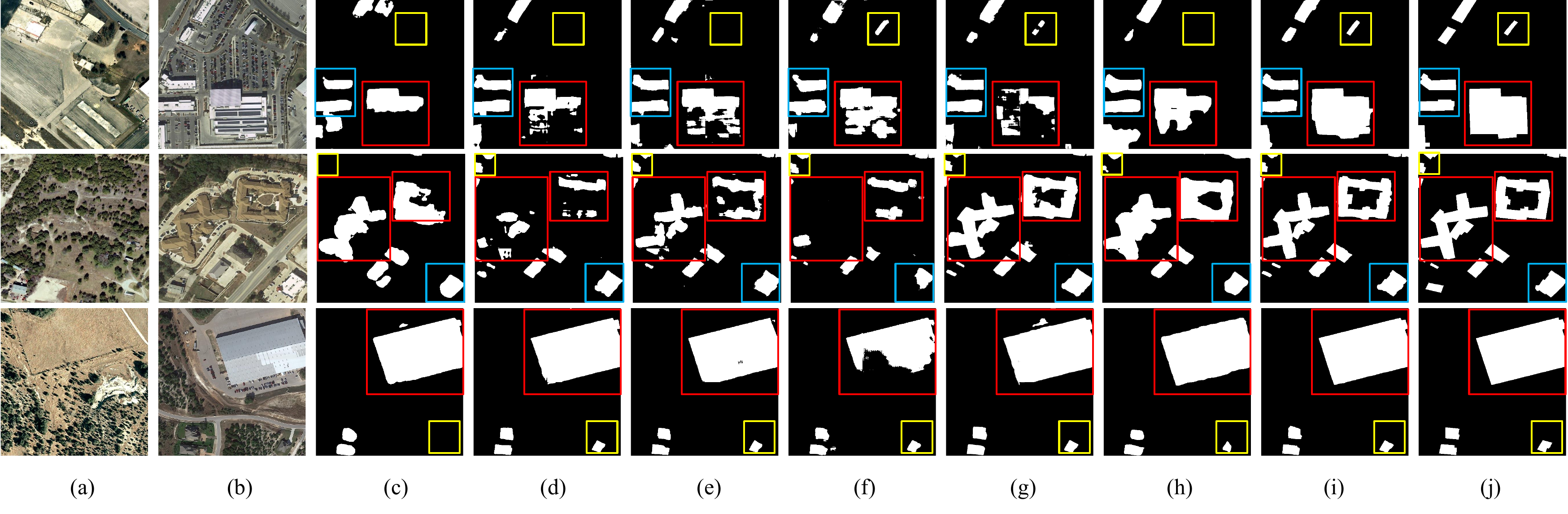}
\end{center}
    \caption{Qualitative results of different CD methods of targets at different scales. (a) $T_1$ image. (b) $T_2$ image. (c) FC-Siam-diff. (d) BiT. (e) ChangeFormer (f) USSFC-Net. (g) SARAS-Net. (h) Changer. (i) FINO(ours). (j) Ground truth.}
    \label{fig:sup1}
\end{figure*}

\textbf{Comparison with other Feature Fusion and Enhancement methods}
To compare the superiority of FINO for feature learning at different scales, we compare it with several methods of recent two years that perform well on the LEVIR-CD dataset, which apply feature concatenation~\cite{22tgars:RSCD}, feature interaction~\cite{tgars:dessn, tgars:darn}, skip connection~\cite{dtnet}, and attention mechanisms~\cite{ChangeFomer} to achieve feature fusion of different scales. 

% \subsection{Comparison with other Feature Fusion and Enhancement methods}
% To compare the superiority of FINO for feature learning at different scales, we compare it with several methods of recent two years that perform well on the LEVIR-CD dataset, which apply feature concatenation~\cite{22tgars:RSCD}, feature interaction~\cite{tgars:dessn, tgars:darn}, skip connection~\cite{dtnet}, and attention mechanisms~\cite{ChangeFomer} to achieve feature fusion of different scales. 

The specific results are shown in the Table~\ref{tab:enhanceEX}. The Table~\ref{tab:enhanceEX} visualizes the considerable improvement of our method over methods such as feature fusion. On both F1 and IoU performing better than DARNet~\cite{tgars:darn} which is the the SOTA, where F1 is 0.43 higher than DARNet and IoU is 0.83 higher than DARNet. 

\begin{figure*}[ht]
\begin{center}
\includegraphics[width=0.95\linewidth]{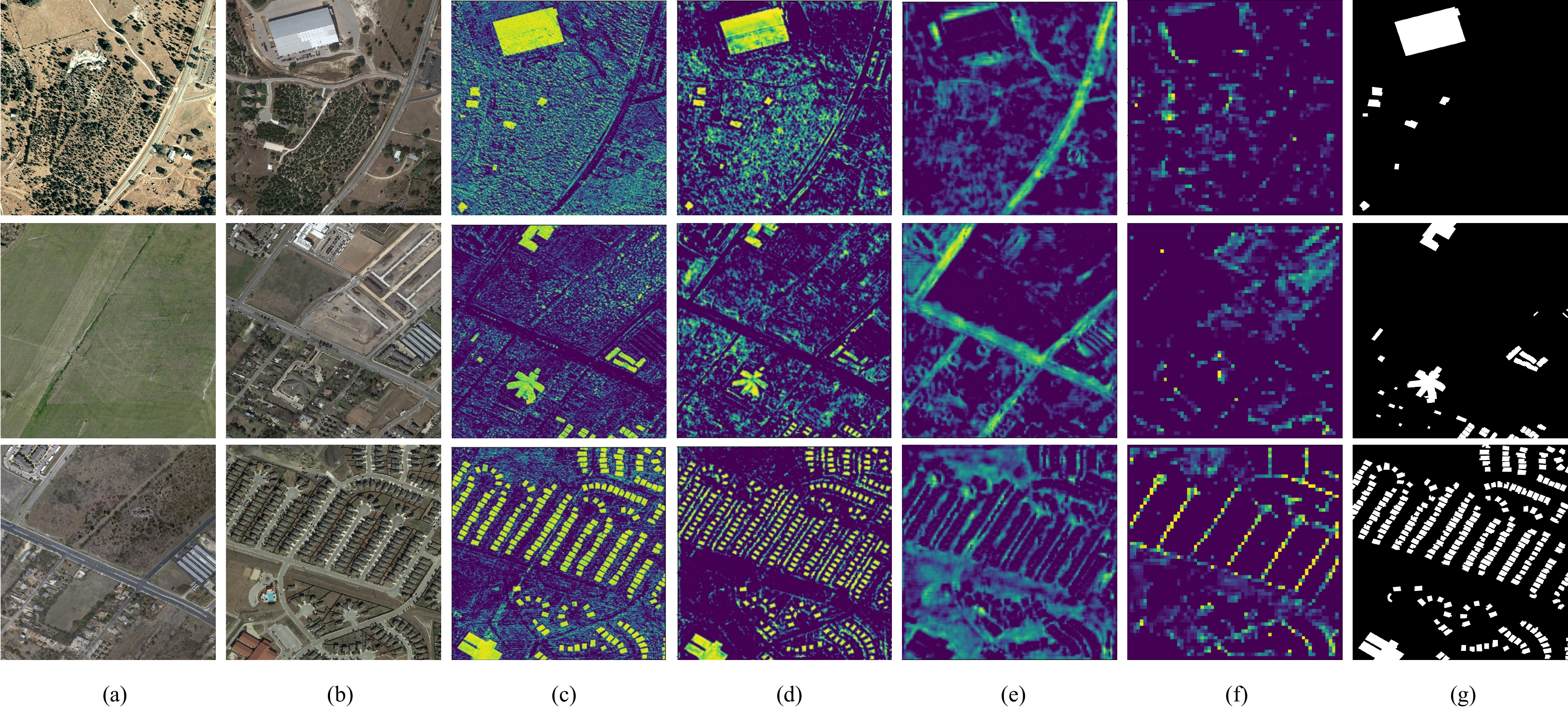}
\end{center}
    \caption{Feature visualisation on LEVIR-CD. (a) $T_1$ image. (b) $T_2$ image. (c) change features $C_1$ of Stage-1. (d) change features $C_2$ of Stage-2. (e) change features $C_3$ of Stage-3. (f) change features $C_4$ of Stage-4. (g) Ground Truth. Yellow pixels represent higher probability of change and blue pixels represent lower probability of unchanged.}
    \label{fig:mid features}
\end{figure*}
\subsection{Detection Result}
To strongly support our methods, we compare the detection effects in different contexts on LEVIR-CD. Overall from the whole images, our method outperforms the other methods. Specifically, as shown in Figure~\ref{fig:sup1}, our method has excellent detection results for both large targets labelled by red boxes, medium targets labelled by blue boxes and small targets labelled by yellow boxes in the same scene.

As shown in Figure~\ref{fig:sup1}, our method also performs well with dense targets. As described in the main text, the context perceptual reconstruction architecture (FINO) architecture is designed to recover the detailed information such as scale and shape. Therefore, our method is more sensitive to the scale and shape of the target. As shown in Figure~\ref{fig:sup1}, the boundaries between the dense targets are very clear in our result, unlike other methods that miss some targets or have fuzzy boundaries. Especially in the regions we have marked with red boxes. Due to the high resolution, please zoom in to view the image if necessary, it will be clearer. 

As shown in Figure~\ref{fig:sup1}, our method also outperforms other methods in complex scenes. Our method is more robust to noise and provides more accurate detection results. The buildings are highly similar to their surroundings, as shown in the first and second rows of the Figure~\ref{fig:sup1}. And the buildings are very similar in shape and colour to the surrounding bush noise, especially as shown in the the red boxes in the second rows and the third rows. Comparing columns (c), (h), and (i), it is clear that our method performs better. Comparing columns (d)-(g) and (i), there are few targets missed. However they detect some noise as a change (e.g. the blue box) because they are oversensitive about the change. Since REGA is designed to eliminate noise and robustly enhance the foreground, our method detects all the changed targets without being disturbed by these noises. This demonstrates the high robustness of our approach.
\subsection{Ablation Study and Further Analysis}
In this section, we provide ablation studies and an analysis of our model. Particularly, we explore the contributions of main components.

\textbf{Importance of the Main Components}
To verify the usefulness of our FINO for information recovery, we conducted ablation experiments for the positions of the cascade structure, where the network layers are numbered 1, 2, 3 and 4 from the lower to the higher levels in Fig.~\ref{fig:framwork}. The specific results are shown in Table~\ref{tab:layer-ex}. As can be seen in Table~\ref{tab:layer-ex}, the accuracy of change detection are greatly improved as the more layers employed FINO. In particular, during the process from Stage-4 to Stage-2, the improvement of FINO is significant.
\begin{table}
\large
    \centering
    \resizebox{0.8\linewidth}{!}{
        \begin{tabular}{c|cccc}
            \toprule[1.5pt]
            stages & Pre.(\%) & Rec.(\%) & F1(\%) & IoU(\%) \\
            \midrule
            4 & 87.43 & 86.84 & 87.84 & 78.31 \\
            4,3 & 92.00 & 89.58 & 90.77 & 83.11\\
            4,3,2 & 93.45 & 90.47 & 91.93 & 85.07\\
            4,3,2,1 & 93.70 & 91.15 &92.41 & 85.89\\
            \bottomrule[1.5pt]
        \end{tabular}
    }
    \caption{A comparative experiment on LEVIR-CD of stages in the cascade structure of our approach, where stages represent the location number of the ResNet network layers in Fig.~\ref{fig:framwork} that used FINO.}
    \label{tab:layer-ex}
    \vspace{-5mm}
\end{table}

To verify the effectiveness of each component, ablation experiments are carried out on CDL, BSA and REGA, as shown in the Table~\ref{tab:Com.ablation}. We use Siam-ResNet as a baseline and verify the performance improvements of the above components on the method in turn. As can be observed from the F1 and IoU in the second row of the Table~\ref{tab:Com.ablation}, the CDL improves the performance of the method enormously that further increase F1 by 2.91\% and IoU by 4.68\%. It can be noticed from the first and third row of the table that the task-specific nois decouple module REGA is able to improve the model performance greatly, making the F1 increase by 4.36\% and IoU by 7.1\%. As shown in the first and forth row of Table~\ref{tab:Com.ablation}, F1 increases 4.66\% and IoU increases 7.61\%. The BSA further enhances our capability to detect changes in our approach, shown in row 5 of Table~\ref{tab:Com.ablation}. 

\textbf{Importance of Fine-Grained Information}
To verify the effectiveness of CDL and proof of significance of fine-grained information, we replace CDL with several other types of attention, including feature concatenation, cross-attention, no-local attention and spatial pyramid pooling (SPP) attention. And due to space constraints, specific structural details are provided in the supplementary material. 

The results on LEVIR-CD are shown in Table~\ref{tab:attention}. Since other attention mechanisms have to rely on matrix multiplication, which has high computational complexity at high resolution, we have to downsample. We downsample the features to a size of $64\times64$. As shown in Table~\ref{tab:attention}, our CDL has a significant superiority over other attention mechanisms, specifically, CDL improves the F1 metric by 1.27\% and IoU by 2.17\% over the Concatenation, which is the best among other attention mechanisms. This further validates the importance of the importance of fine-grained information in change detection.
\begin{table}[tb]
    \centering
    \setlength{\tabcolsep}{1.0mm}{
    \resizebox{\linewidth}{!}{
    \begin{tabular}[]{c|ccc|cccc}
        \toprule[1.5pt]
        Row & CDL & BSA & REGA & Pre.(\%) & Rec.(\%) & F1(\%) & IoU(\%) \\
        \midrule
        1 & & & & 87.26 & 86.70&  87.02 & 77.03 \\
        2 &$\surd$ & & & 90.14 & 89.73 & 89.93 & 81.71  \\
        3 &  & $\surd$ & & 92.22 & 90.56 & 91.38 & 84.13  \\
        4 &$\surd$ & $\surd$ & & 93.14 & 90.27 & 91.68 & 84.64  \\
        5 &$\surd$ &$\surd$ & $\surd$ & 93.70 & 91.15 & 92.41 & 85.89\\
        \bottomrule[1.5pt]
    \end{tabular}
    }
    }
    \caption{Object change detection experimental results on LEVIR-CD for validation of the effectiveness of each component, where Sup. denotes the indirect intermediate supervision of learnable mask.}
    \label{tab:Com.ablation}
    \vspace{-3mm}
\end{table}
\begin{table}[t]
    \centering
    \begin{tabular}[width=\linewidth]{c|cccc}
    \toprule[1.5pt]
        Replace & Pre.(\%) & Rec.(\%) & F1(\%) & IoU(\%) \\ \midrule
        Concatenation &92.53&89.79&91.14&83.72\\
        Cross-attention&88.85 &81.37 &84.94 & 73.82 \\
        No-local attention&90.80&88.64 &89.75 &81.41  \\
        SPP attention&91.97 &89.44 & 90.69& 82.96\\
        CDL &93.70&91.15 &92.41 &85.89\\
        \bottomrule[1.5pt]
    \end{tabular}
    \caption{Comparison of CD performance on LEVIR-CD dataset by using our proposed CDL and other attention methods.}
    \label{tab:attention}
    \vspace{-5mm}
\end{table}

\textbf{Visualization of FINO}
In order to objectively show whether the fine-grained information is adequately compensated by FINO, we visualize the difference features of the four stages in FINO. It is shown in Fig.~\ref{fig:mid features}. In the feature maps of column (c)-(d) in Fig.~\ref{fig:mid features}, the pixels around the yellow areas appear darker and have a clear contrast with the surrounding pixels, with details of objects clearly. It proves that FINO is effective in noise elimination in non-changing regions and can effectively recover detailed texture information of images. In the feature maps column (e)-(f) of Fig.~\ref{fig:mid features}, no details such as the shape of objects can be distinguished, but the outline is clear. This proves that as the network layer deepens, showing in column (e)-(f) of Fig.~\ref{fig:mid features}, the high-level semantic features of object are very obvious and less noisy, but lost a lot of detail informations. And due to FINO, the low-level change features of Fig.~\ref{fig:mid features} (c) has few noise, including the higher-level semantics of the change features retained, which is used to predict changes. And comparing with the GT of Fig.~\ref{fig:mid features} (g), all the yellow region in the features of Fig.~\ref{fig:mid features} (c) are the change objects.
\section{Conclusion}
In this article, we propose the fine-grained information compensation and noise decoupling (FINO) architecture to address the issues of information loss in convolutional networks and the interference of pseudo-change noise in change detection. Our designed context-aware learning (CDL) fully compensates for fine-grained information. We introduce a brightness-aware and shape-aware module to mitigate task-agnostic noise, and propose a regularization gate (REGA) structure to decouple task-specific noise. Extensive experiments demonstrate the advantages of our FINO, outperforming current state-of-the-art methods on mutiple change detection datasets.

%%
%% The acknowledgments section is defined using the "acks" environment
%% (and NOT an unnumbered section). This ensures the proper
%% identification of the section in the article metadata, and the
%% consistent spelling of the heading.
% \begin{acks}
% To Robert, for the bagels and explaining CMYK and color spaces.
% \end{acks}

%%
%% The next two lines define the bibliography style to be used, and
%% the bibliography file.
\bibliographystyle{ACM-Reference-Format}
\bibliography{sample-base}

\end{document}